%% file: main.tex
\documentclass[10pt,twocolumn,letterpaper]{article}

\usepackage{wacv}              


\usepackage[nolist]{acronym}
\usepackage{amsmath,amssymb,amsfonts}
\usepackage{bbold}
\usepackage{subcaption}

\definecolor{wacvblue}{rgb}{0.21,0.49,0.74}
\usepackage[pagebackref,breaklinks,colorlinks,allcolors=wacvblue]{hyperref}
\usepackage[accsupp]{axessibility}

\begin{acronym}
\acro{al}[AL]{active learning}
\acro{cnn}[CNN]{convolutional neural network}
\acro{dcn}[DCN]{deformable convolution}
\acro{dice}[DSC]{Dice similarity coefficients}
\acro{ema}[EMA]{exponential moving average}
\acro{froc}[FROC]{Free Response Operating Characteristic}
\acro{gnn}[GNN]{graph neural network}
\acro{he}[H\&E]{hematoxylin \& eosin}
\acro{hact}[HACT]{HierArchical cell-to-tissue}
\acro{iou}[IoU]{intersection over union}
\acro{miou}[mIoU]{class-averaged intersection over union}
\acro{roi}[RoI]{region of interest}
\acro{wsi}[WSI]{whole slide image}
\acro{udh}[UDH]{usual ductal hyperplasia}
\acro{adh}[ADH]{atypical ductal hyperplasia}
\acro{fea}[FEA]{flat epithelial atypia}
\acro{dcis}[DCIS]{ductal carcinoma in situ}
\end{acronym}

\def\wacvPaperID{1994} 
\def\confName{WACV}
\def\confYear{2026}

\def\MYTITLE{Decomposition Sampling for Efficient Region Annotations in Active Learning}
\title{\MYTITLE}

\def\MYAUTHOR{
    Jingna Qiu\textsuperscript{1,2} \quad
    Frauke Wilm\textsuperscript{1,3} \quad
    Mathias Öttl\textsuperscript{1,3} \quad
    Jonas Utz\textsuperscript{1} \quad \\
    Maja Schlereth\textsuperscript{1} \quad
    Moritz Schillinger\textsuperscript{1} \quad
    Marc Aubreville\textsuperscript{4}  \quad
    Katharina Breininger\textsuperscript{2} \\
    \textsuperscript{1}Friedrich-Alexander-Universität Erlangen-Nürnberg
    \textsuperscript{2}Julius-Maximilians-Universität Würzburg\\
    \textsuperscript{3}MIRA Vision Microscopy GmbH
    \textsuperscript{4}Hochschule Flensburg\\
    {\tt\small {jingna.qiu@fau.de \quad katharina.breininger@uni-wuerzburg.de}}
}
\author{\MYAUTHOR}

\begin{document}
\maketitle
\input{sec/0_abstract}

\input{sec/1_intro}
\input{sec/2_related_work}
\input{sec/3_approach}

\input{sec/4_experiments}

\input{sec/5_conclusion}
\input{sec/7_acknowledgement}
{
    \small
    \bibliographystyle{ieeenat_fullname}
    \bibliography{main}
}

\clearpage
\newpage 
\setcounter{page}{1}
\setcounter{section}{0}

\twocolumn[
\begin{center}
    {\Large \bfseries{\MYTITLE: Supplementary Materials}}\\
    \author{\MYAUTHOR}
\end{center}
]

\input{sec/6_supplementary}  
\end{document}

%% file: sec/0_abstract.tex
\begin{abstract}
Active learning improves annotation efficiency by selecting the most informative samples for annotation and model training. While most prior work has focused on selecting informative images for classification tasks, we investigate the more challenging setting of dense prediction, where annotations are more costly and time-intensive, especially in medical imaging. Region-level annotation has been shown to be more efficient than image-level annotation for these tasks. However, existing methods for representative annotation region selection suffer from high computational and memory costs, irrelevant region choices, and heavy reliance on uncertainty sampling. We propose \textbf{decomposition sampling (DECOMP)}, a new active learning sampling strategy that addresses these limitations. It enhances annotation diversity by decomposing images into class-specific components using pseudo-labels and sampling regions from each class. Class-wise predictive confidence further guides the sampling process, ensuring that difficult classes receive additional annotations. Across ROI classification, 2-D segmentation, and 3-D segmentation, DECOMP consistently surpasses baseline methods by better sampling minority-class regions and boosting performance on these challenging classes. Code is in \url{https://github.com/JingnaQiu/DECOMP.git}.
\end{abstract}

%% file: sec/1_intro.tex
\section{Introduction}
\label{sec:intro}

Dense prediction tasks such as segmentation demand extensive high-quality annotations. In medical imaging, annotation is particularly challenging due to ambiguous inter-class tissue boundaries and substantial intra-class morphological variation, often leading to disagreement among experts~\cite{litjens20181399, wilm2022pan}. To mitigate these challenges, strategies such as refining labeling protocols~\cite{radsch2023labelling}, multi-annotator review~\cite{heller2019kits19}, and model-assisted pre-annotation~\cite{bertram2019large} have been explored. Another promising direction is \ac{al}~\cite{settles2009active, budd2021survey}, which improves efficiency by selectively annotating only informative samples while maintaining full-annotation performance. By reducing the annotation amount, \ac{al} enables experts to focus on fewer but more valuable cases, thereby improving label quality.

\ac{al} operates iteratively. Starting with an initial labeled set (typically selected at random), a baseline model is trained. In each subsequent cycle, new samples are chosen based on the current model with a defined sampling strategy. These samples are manually annotated, added to the labeled set, and used to update the model. The process continues until performance goals are met or the annotation budget is exhausted. Among sampling strategies, \textit{uncertainty sampling} targets samples where the model is least confident, while \textit{diversity sampling} prioritizes representative samples that capture dataset variability and reduce redundancy.  

Originally developed for identifying informative images in classification tasks, \ac{al} has been extended to region-based annotation for dense annotation tasks like segmentation. For example, annotating $128 \times 128$ regions was found more efficient than labeling entire $1024 \times 2048$ images on Cityscapes~\cite{mackowiak2018cereals}. More recently, region-based \ac{al} has also been applied to object detection~\cite{vo2022active}, optical flow prediction~\cite{yuan2022optical}, and autonomous driving perception~\cite{segal2022just}. In large-scale medical imaging, where 3-D scans contain hundreds of slices and histopathological slides reach gigapixel sizes, region-based \ac{al} is especially promising for distributing annotation budgets more effectively.

This paper focuses on diversity sampling for region-based \ac{al}. Standard approaches, adapted from image selection, include clustering-based~\cite{nguyen2004active} and core-set-based~\cite{sener2017active} methods. Both require computing and storing features for all candidate regions, then clustering selects cluster centers, while core-set greedily identifies a subset that maximally covers the feature space. These methods, however, have notable limitations. First, computing and storing features for all regions is computationally and memory intensive. For example, a single $1024 \times 2048$ image yields 128 non-overlapping $128 \times 128$ regions, with complexity increasing dramatically for 3-D volumes or gigapixel medical images. This effect is even more pronounced if overlapping regions are considered. Second, these methods are task-agnostic, and thus may select irrelevant regions with limited training value for the target task. Third, they treat easy and hard samples equally, ignoring current model weaknesses. Consequently, diversity sampling is often combined with uncertainty sampling, which pre-filters candidates~\cite{yang2017suggestive,jin2021reducing}. While this improves relevance and reduces computational cost, it compromises diversity because uncertain samples can be highly similar.

To address these limitations, we propose \textbf{decomposition sampling (DECOMP)}, a new strategy for selecting diverse annotation regions. DECOMP decomposes each image into class-specific components using pseudo-labels and samples regions across classes to ensure diversity. It further incorporates class-wise model confidence to prioritize uncertain classes and address model weaknesses. Unlike conventional methods, DECOMP (i) requires only a single forward prediction pass of the entire image without costly feature storage or recomputation for overlapping regions, (ii) ensures task relevance by selecting from predicted classes (particularly advantageous for small target structures within large backgrounds, such as kidney tumors in abdominal CT scans), and (iii) directly integrates class-wise confidence estimates to identify difficult cases without relying on separate uncertainty filtering.

We evaluate DECOMP on three tasks: \ac{roi} classification in histopathology (BRACS~\cite{brancati2022bracs}), 2-D segmentation in natural scenes (Cityscapes~\cite{cordts2016cityscapes}), and 3-D segmentation in CT scans (KiTS23~\cite{heller2019kits19}). The histopathology \ac{roi} classification task reflects an annotation process of coarse segmentation, completely identifying and categorizing \acp{roi} across multiple classes in gigapixel \acp{wsi} requires substantial expert annotation effort. Cityscapes is included to assess generalizability beyond medical imaging. Results demonstrate that DECOMP consistently improves annotation efficiency compared to state-of-the-art baselines. Our main contributions are:  
\begin{itemize}
    \item We introduce DECOMP, a diversity-based \ac{al} strategy that selects annotation regions via class decomposition. 
    \item We address limitations of conventional methods, including high computational cost, irrelevant region selection, and reliance on uncertainty sampling.
    \item We perform extensive evaluation across \ac{roi} classification, 2-D and 3-D segmentation, proving that DECOMP achieves competitive performance with substantially fewer annotations compared to strong baselines. 
\end{itemize}

\begin{figure*}[t]
{\centering\centerline{\includegraphics[width=.8\paperwidth]{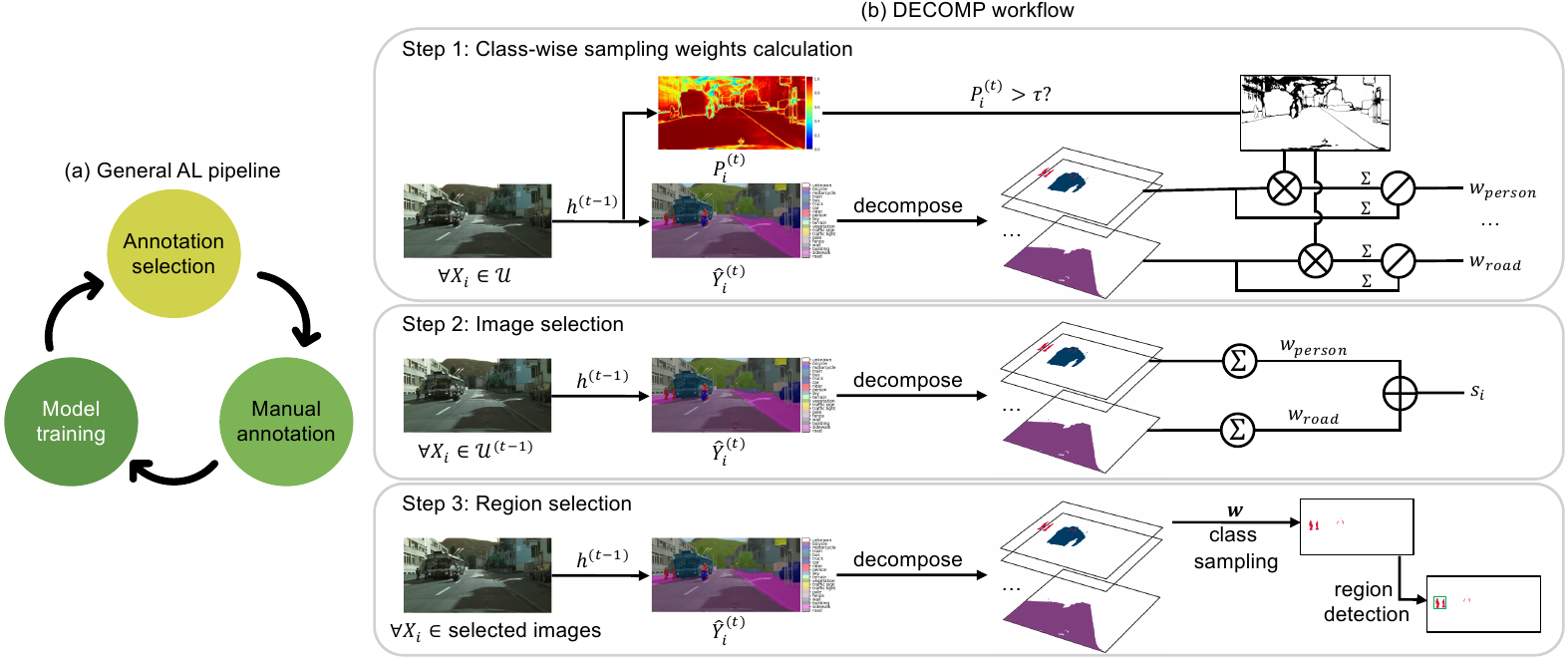}}}
\caption{(a) \Ac{al} builds the annotation set and model iteratively. The last trained model is used to select new annotation samples, which are then labeled and added to the existing annotations for a new round of model update. (b) DECOMP enhances annotation diversity by querying annotations across different classes. It first segments an unlabeled image $X_i$ into class components using predictions from the model trained in the previous \ac{al} cycle $h^{(t-1)}$. Class-wise sampling weights $\mathbf{w}$ are computed based on the proportion of high-confidence class predictions to prioritize annotations for uncertain classes. To select the most informative image, a score $s_i$ is determined for each unlabeled image by summing its predicted class frequencies, weighted by the sampling weights. For each selected image, regions are sequentially identified by first sampling a class based on $\mathbf{w}$ and then selecting the region that best represents it.}
\label{fig:method}
\end{figure*}

%% file: sec/2_related_work.tex
\section{Related Work}
\label{sec:related_work}

\noindent\textbf{Active Learning}
\ac{al} improves annotation efficiency by selecting samples expected to yield the greatest performance gains, typically through uncertainty or diversity measures. 

Uncertainty sampling prioritizes samples for which the model is least confident, using least confidence-, margin-, or entropy-based criteria~\cite{settles2009active}. Other approaches approximate predictive variance using model ensembles~\cite{beluch2018power} or Monte Carlo dropout~\cite{gal2017deep}. Gradient-based methods such as BADGE~\cite{ash2019deep} select samples with large loss gradients, while loss-prediction approaches estimate which samples will incur high training loss~\cite{yoo2019learning}.

Diversity-based methods reduce redundancy by selecting representative samples, commonly via clustering~\cite{nguyen2004active} or core-set selection~\cite{sener2017active}, often combined with uncertainty filtering~\cite{yang2017suggestive,jin2021reducing}. Refinements include estimating cluster numbers via dendrograms or defining core-sets per cluster~\cite{zheng2019biomedical}, as well as applying clustering to gradient embeddings (BADGE)~\cite{ash2019deep}. Identifying samples from high-density regions in feature space has also been explored~\cite{jin2023density}. 

In line with recent systematic evaluations~\cite{luth2023navigating}, we consider entropy and least confidence as uncertainty baselines; clustering and core-set selection as diversity baselines; and BADGE as a combined approach.

\noindent\textbf{Class-aware Training and Sampling}
DECOMP’s use of class-wise confidence is inspired by FlexMatch~\cite{zhang2021flexmatch}, which filters pseudo-labels during training based on confidence; in contrast, DECOMP uses confidence to guide annotation selection. Unlike class-aware methods in imbalanced learning, which balance datasets via oversampling or re-weighting, DECOMP addresses imbalance at the annotation stage. Class confidence has also been used for image selection in \ac{al}: Hu \etal~\cite{hu2023learning} weight image-level class probabilities by the inverse of class confidence to identify difficult-to-learn annotation images for classification tasks, and Qiu \etal~\cite{qiu2023class} sum class-presence frequencies weighted by inverse confidence for segmentation tasks. DECOMP follows a similar image-selection idea but additionally applies confidence-aware weighting to region selection.

%% file: sec/3_approach.tex
\section{Approach}
\subsection{General Setup and Notation}

As illustrated in \cref{fig:method}(a), region-based \ac{al} proceeds in iterative cycles of region selection, manual annotation, and model retraining. Given a pool of unlabeled images $\mathcal{U}=\{X_{1}, \dots, X_{N}\}$, we initialize with $\mathcal{U}^{(0)}=\mathcal{U}$ and an empty labeled set $\mathcal{L}^{(0)}=\varnothing$. In the first cycle, regions are randomly sampled to form the initial labeled set $\mathcal{L}^{(1)}$, which trains a baseline model $h^{(1)}$ and updates the unlabeled set as $\mathcal{U}^{(1)}=\mathcal{U}^{(0)}\setminus\mathcal{L}^{(1)}$. At each subsequent cycle $t$, regions are selected using predictions from $h^{(t-1)}$, annotated by humans, and added to form $\mathcal{L}^{(t)}$ to train a new model $h^{(t)}$. This process repeats until a performance target or annotation budget is reached.

Each image $X_{i}\in\mathbb{R}^{H_{i}\times W_{i}}$ begins fully unlabeled, with regions progressively selected and annotated. In each \ac{al} cycle, $n_{\text{image}}$ images are chosen, and $n_{\text{region}}$ regions are annotated per selected image. Following community practice~\cite{mackowiak2018cereals}, each region is defined as an $l\times l$ square. The ground-truth annotation for the $j^{\text{th}}$ region $r_{ij}$ is denoted as $y(r_{ij})\in\{1,\dots,C\}^{l\times l}$, where $C$ is the number of classes; unselected areas remain unlabeled. Predictions for the entire image $X_{i}$ from $h^{(t-1)}$ are represented as pseudo-labels $\hat{Y}_{i}^{(t)}\in\{1,\dots,C\}^{H_{i}\times W_{i}}$ with class probabilities $P_{i}^{(t)}\in[0,1]^{H_{i}\times W_{i}}$, retaining only the maximum probability per pixel. For 3-D images, we select 2-D annotation regions rather than 3-D cubes, reflecting slice-by-slice annotation workflows~\cite{heller2023kits21} and improving spatial coverage and tissue variability while reducing redundancy from adjacent slices.

This notation extends naturally to other dense annotation tasks such as \ac{roi} classification. In this case, each image $X_i$ contains $M$ manually defined \acp{roi}, $\{r_{ij}|_{j=1}^M\}$. These serve directly as candidate regions. Model predictions are given as $\hat Y_{i}^{(t)}=\{\hat y^{(t)}(r_{ij})\in\{1, ..., C\}|_{j=1}^M\}$ with associated probabilities $P_{i}^{(t)}=\{p^{(t)}(r_{ij})\in[0, 1]|_{j=1}^M\}$.

\subsection{Decomposition Sampling (DECOMP)}
\Cref{fig:method}(b) shows our proposed method DECOMP. At its core, DECOMP decomposes each unlabeled image into $C$ class-specific components using predictions $\hat{Y}_i^{(t)}$ from the model developed in previous cycle $h^{(t-1)}$. This provides two benefits:  
(1) by explicitly identifying regions predicted as each class, it ensures balanced coverage across classes in annotation, including minority categories (\eg rare disease patterns); and  
(2) it enables class-wise confidence estimation, which we use to derive sampling weights that prioritize difficult classes. We now describe the computation of class-wise sampling weights and their role in annotation selection. For clarity, the cycle index $t$ is omitted.

\subsubsection{Class-wise Sampling Weights}
Sampling weights $\mathbf{w} = (w_1, \dots, w_C)$ modulate annotation distribution across classes. Intuitively, classes with low predictive confidence are prioritized, so that annotations provide greater training benefit.

Inspired by FlexMatch~\cite{zhang2021flexmatch}, class confidence $\sigma_c$ for class $c$ is estimated as the fraction of high-confidence predictions among all predictions of class $c$, as
\begin{equation}
    \sigma_c = \frac{\sum_{i \in \mathcal{U}} \sum_{j \in X_i} \mathbb{1}(P_i(j) > \tau \wedge \hat{Y}_i(j)=c)}{\sum_{i \in \mathcal{U}} \sum_{j \in X_i} \mathbb{1}(\hat{Y}_i(j)=c)},
    \label{Eq:tau}
\end{equation}
where $\tau$ is a confidence threshold and $j$ denotes pixels (for segmentation) or \acp{roi} (for \ac{roi} classification). Prior work shows that $\tau$ is task-dependent~\cite{zhang2021flexmatch,hu2023learning}; we empirically set $\tau=0.7$ and analyze its effect in ablation studies. Sampling weights are then defined, as
\begin{equation}
    w_c = \frac{1 - \sigma_c}{\sum_{c'=1}^C (1 - \sigma_{c'})}.
\end{equation}
With this formulation, uncertain classes receive higher sampling weights to be sampled more often. Weights are updated at the beginning of each cycle to adapt to evolving model weaknesses, and can be further manually adjusted to emphasize clinically important or rare classes.

\subsubsection{Image Selection}

Prior work typically selects regions across the full unlabeled pool $\mathcal{U}$ to maximize diversity~\cite{mackowiak2018cereals}. In medical imaging, however, annotators must review entire-image context before annotation, making this strategy time-intensive. Instead, we propose annotating regions densely within a smaller set of images, reducing cognitive load and annotation time. Since unannotated images are unused in training, this also lowers memory cost. We validate the effectiveness of this strategy in ablation studies by comparing dense annotation of fewer images with sparse annotation of more images. To identify the most informative images, we score each unlabeled image $X_i$ by weighting its predicted class frequencies with $\mathbf{w}$, as
\begin{equation}
    s_i = \sum_{c=1}^C w_c \cdot \big(\sum_{j \in X_i} \mathbb{1}(\hat{Y}_i(j)=c)\big),
\end{equation}
In this way, images rich in uncertain classes receive higher scores and are prioritized for annotation. To prevent large classes (\eg sky or road in Cityscapes) from dominating image scores due to their high pixel counts, we cap predicted class frequencies before weighting: at 10\% of the image size for segmentation and at 1 ROI for ROI classification. This ensures that sampling is driven by class uncertainty rather than class size.

\subsubsection{Region Selection}
\label{sec:region_selection}
Within each selected image, regions are sequentially chosen to maximize informativeness and diversity:
1) Sample a class $c$ via random sampling based on probabilities $\mathbf{w}$, prioritizing uncertain but also ensuring representation across all classes. This prevents excessive focus on noisy or inherently difficult classes. 
2) Identify the region best representing $c$. For segmentation, an $l \times l$ sliding window scans $\hat{Y}_i$, and the region containing the largest predicted area of $c$ is selected, excluding any overlap with already selected regions; integral images are used for efficient computation. For 3-D volumes, a slice containing $c$ is randomly chosen before selecting a 2-D region. For \ac{roi} classification, the \ac{roi} with the highest probability of $c$ is selected.

Unlike clustering- or core-set–based diversity methods, which require repeated feature extraction and storage and become prohibitive for large medical images, especially when considering overlapping regions, DECOMP needs only a single forward prediction per image, greatly simplifying deployment. Its computational complexity also remains independent of the number of classes, as class decomposition and weight estimation are derived directly from this single prediction, image scoring involves simple class-frequency counting, and region selection is performed $n_{\text{region}}$ times. Moreover, DECOMP accommodates selecting regions of arbitrary shape without additional overhead.

%% file: sec/4_experiments.tex
\section{Experiments}
\subsection{Experimental Setups}

\noindent\textbf{Datasets.} We evaluate on three benchmarks. \textbf{BRACS}~\cite{brancati2022bracs} contains $320$ \ac{he}-stained \acp{wsi} (train/val/test: $193/68/59$). Annotating a \ac{wsi} requires complete \ac{roi} identification and classification into benign, atypical, or malignant breast cancer lesions. The splits include $3,163/602/626$ \acp{roi}, with $1$–$119$ \acp{roi} per slide (median: $8$). Test set evaluation uses pre-extracted \acp{roi}. \textbf{Cityscapes}~\cite{cordts2016cityscapes} provides $2,975$ training and $500$ validation street-scene images for 2-D segmentation across $19$ classes. \textbf{KiTS23}~\cite{heller2023kits21} comprises $489$ 3-D CT scans ($60$–$610$ slices; median: $177$) for hierarchical kidney, tumor-and-cyst, and tumor segmentation. All datasets are fully annotated, but labels for unselected regions are hidden following standard \ac{al} practice.

\noindent\textbf{Active learning setups.}  
Each dataset highlights different annotation challenges, enabling a comprehensive evaluation of DECOMP. To ensure robustness, we evaluate performance under multiple per-cycle annotation budgets, as this factor strongly influences \ac{al} effectiveness~\cite{luth2023navigating,qiu2023adaptive,mackowiak2018cereals}. For BRACS, we tested four settings combining $n_{\text{image}} \in \{1, 5\}$ and $n_{\text{region}} \in \{15, 30\}$, capped at the number of available \acp{roi} for each \ac{wsi}. For Cityscapes, we used $128\times128$ regions, which has been shown to outperform alternative sizes~\cite{mackowiak2018cereals}. We evaluated three settings: $(n_{\text{image}}=25, n_{\text{region}}=10)$, $(n_{\text{image}}=50, n_{\text{region}}=10)$, and $(n_{\text{image}}=100, n_{\text{region}}=5)$. The first two examine the effect of frequent model updates, while the latter two maintain an equal annotation budget per cycle, allowing us to compare dense annotation of fewer images against sparse annotation of more images. For KiTS23, we considered two scenarios: $(n_{\text{image}}=10, n_{\text{region}}=10)$ and $(n_{\text{image}}=20, n_{\text{region}}=10)$, with a region size of $32 \times 32$ pixels (at a spacing of $1.84$ mm, corresponding to a physical FOV of $59 \times 59$ mm$^2$), slightly larger than average kidney tumor size.

\noindent\textbf{full-annotation benchmarks, implementation details, and evaluation metrics.}  
For each dataset, we established full-annotation benchmarks using state-of-the-art methods. To ensure reproducibility and reduce computational costs, we retained the original hyperparameters rather than re-tuning for each \ac{al} cycle with varying annotation sizes, and used the same evaluation metrics on the original evaluation sets. For BRACS \ac{roi} classification, we used HACT-Net~\cite{pati2022hierarchical}, the model developed by the dataset provider, achieving a weighted F1 of $0.7320$ on the test set (mean of five runs). For Cityscapes 2-D segmentation, we used InternImage~\cite{wang2023internimage} with UperNet~\cite{xiao2018unified} as the decoder, owing to its top leaderboard performance\footnote{\url{https://www.cityscapes-dataset.com/benchmarks/}}; to balance accuracy and efficiency under iterative \ac{al}, we adopted the InternImage-T variant. Full-annotation performance reached \ac{miou} $0.8096$ on the validation set (mean of five runs). For KiTS23 3-D segmentation, we used 3-D nnU-Net~\cite{isensee2021nnu}, the most widely adopted method among top KiTS21 leaderboard entries~\cite{heller2021state}, achieving class-averaged \ac{dice} $0.8751$ (mean of five-fold cross-validation). All experiments used one NVIDIA A100 Tensor Core GPU.

\noindent\textbf{Comparison methods.}  
We evaluate against several established \ac{al} strategies. \textbf{RAND} performs random sampling for both image selection and the subsequent selection of non-overlapping regions within the selected images. \textbf{UNCERT} applies uncertainty sampling (entropy for models with a softmax activation function and least confidence for sigmoid models), choosing images with the highest average predictive uncertainty and then selecting the most uncertain regions. In \textbf{DIVERS}, uncertainty sampling is first used to form a candidate pool three times larger than $n_\text{region}$ (3 is an empirical factor informed by prior work using values between 2 and 10~\cite{yang2017suggestive,jin2021reducing}). Representative regions are then chosen from this pool using either clustering~\cite{nguyen2004active} (\textbf{DIVERS(cluster)}) or core-set selection~\cite{sener2017active} (\textbf{DIVERS(core-set)}). \textbf{BADGE}~\cite{ash2019deep} selects informative samples by constructing gradient embeddings for each region and clustering them to promote both uncertainty and diversity. Like DIVERS, BADGE first applies uncertainty sampling to filter a candidate pool. Following~\cite{mackowiak2018cereals}, evaluation is based on 95\% of the full-annotation benchmarks. On BRACS and Cityscapes, all methods start from the same initial labeled set, which varies across repetitions. On KiTS23, each fold uses its own initial labeled set. To maximize the use of the entire unlabeled dataset, all methods select images until all images in $\mathcal{U}$ were sampled, restarting the process in a new loop.

\subsection{Comparison with Baseline Methods}
\begin{figure}[t]
{\centering\centerline{\includegraphics[width=1\columnwidth]{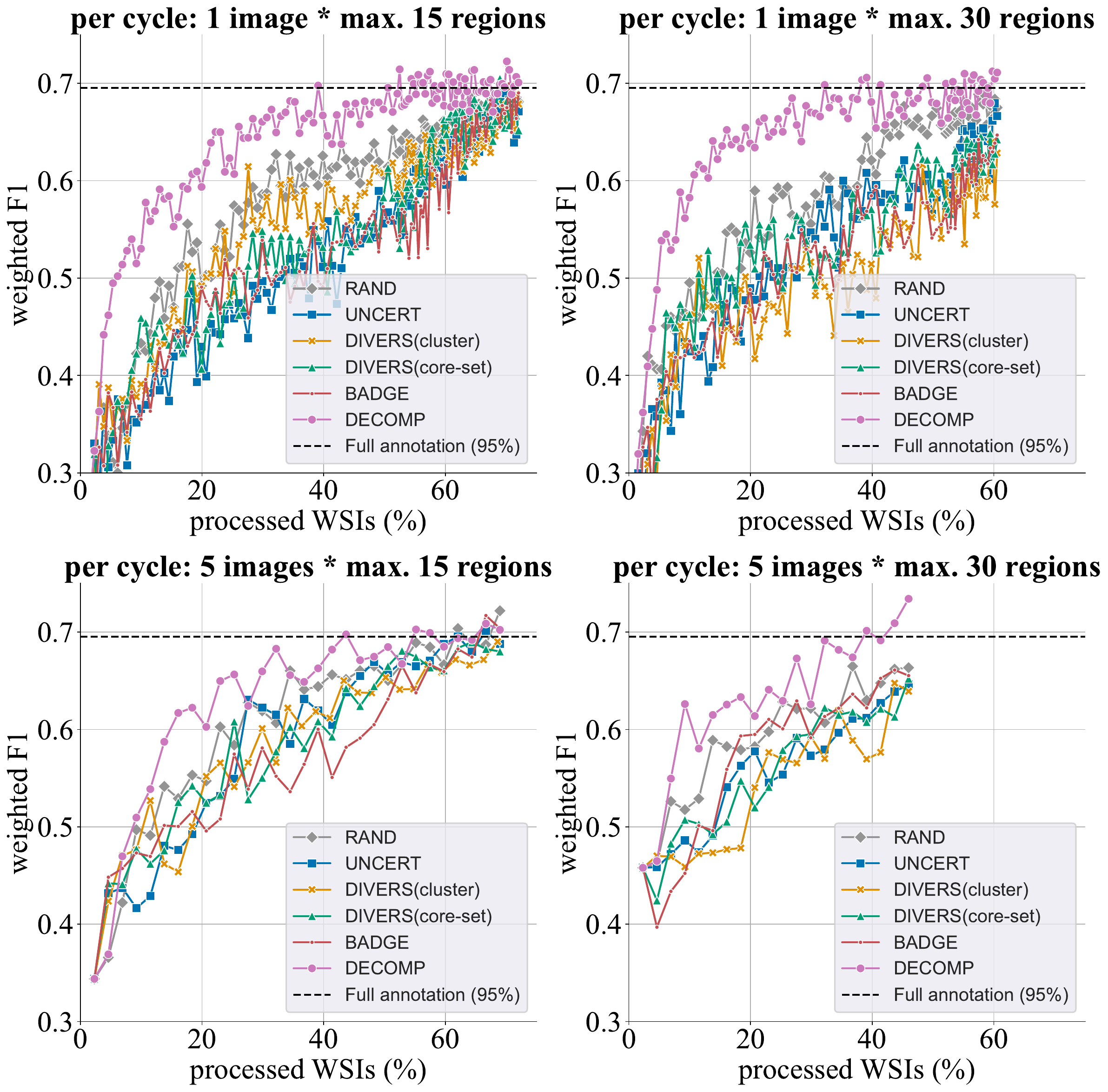}}}
\caption{Results on BRACS test set across $120, 90, 30, 20$ cycles for per-cycle annotation budget combinations $n_\text{image}\in\{1, 1, 5, 5\}$ and $n_\text{region}\in\{15, 15, 30, 30\}$, respectively. Weighted F1 as a function of processed WSIs (\%) for different methods. Scatter points are denser in later cycles because all $68$ validation \acp{wsi} have been fully annotated, leaving only training \acp{wsi} in $\mathcal{U}$. Means over five runs.}
\label{fig:BRACS_results}
\end{figure}

\noindent\textbf{Results on BRACS \ac{roi} classification} Results under different per-cycle annotation budgets are shown in~\cref{fig:BRACS_results}. The goal is to minimize the number of \acp{wsi} requiring expert review (\ie image inspection, \ac{roi} identification, and classification).  Interestingly, RAND outperforms UNCERT under small budgets ($n_\text{image}=1, n_\text{region}=15$), likely because the most uncertain images contain many similar uncertain \acp{roi}, limiting diversity and destabilizing training. UNCERT improves when more regions or more images are selected per cycle. DIVERS and BADGE, which incorporate diversity, do not consistently improve UNCERT, partly because many selected images contain fewer than $n_\text{region}$ \acp{roi}, causing all available regions to be selected regardless of strategy. This limitation also affects DECOMP, yet DECOMP still outperforms DIVERS and BADGE in all settings, indicating stronger image-level selection. Overall, DECOMP reaches 95\% of the full-annotation benchmark after processing only $40\%$ of WSIs, meaning fewer than $105$ \acp{wsi} require review, with at most $15$ \acp{roi} annotated per image.

 \begin{figure}[t]
{\centering\centerline{\includegraphics[width=1\linewidth]{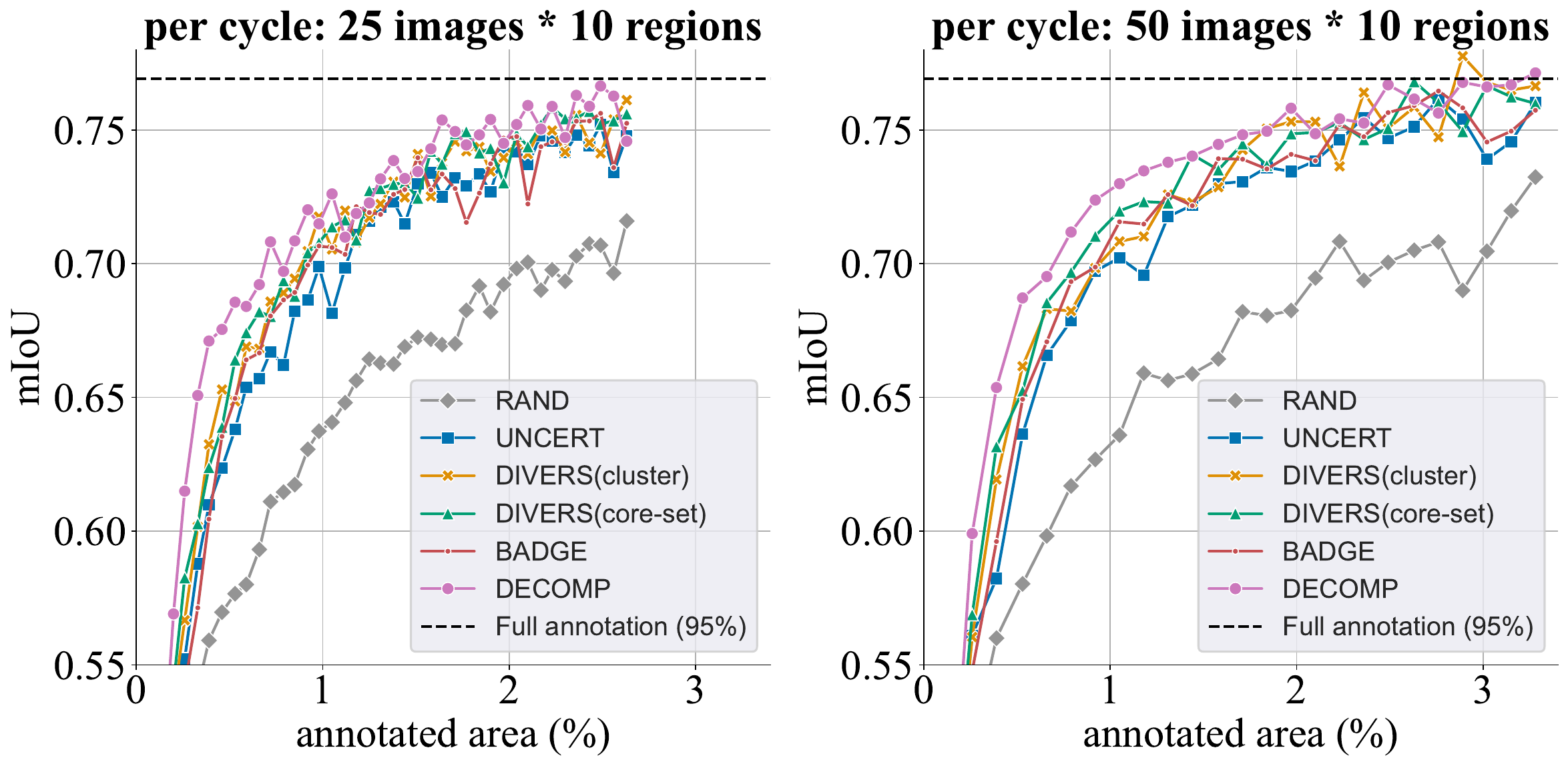}}}
\caption{Results on Cityscapes validation set across $40$ and $25$ cycles for \ac{al} budget combinations $n_\text{image}\in\{25, 50\}$ and $n_\text{region}=10$, respectively. \ac{miou} as a function of annotated area (\%) for different methods. Means over three runs.}
\label{fig:Cityscapes_results}
\end{figure}
\noindent\textbf{Results on Cityscapes 2-D segmentation} Results are shown in~\cref{fig:Cityscapes_results}. Here the goal is to reduce the area requiring time-intensive pixel-level annotation. RAND shows low efficiency, likely due to the extreme class imbalance on Cityscapes (\eg motorcycles cover only $0.27\%$ as many pixels as roads in full-annotation). DIVERS(cluster), DIVERS(core-set), and BADGE modestly improve over UNCERT by selecting representative regions. DECOMP outperforms all baselines, achieving $95\%$ of full-annotation performance with less than $2.5\%$ of the total image area annotated ($9,500$ regions of $128\times128$ pixels from $2,975$ images of $1024\times2048$ pixels each). 


\begin{figure}
{\centering\centerline{\includegraphics[width=1\columnwidth]{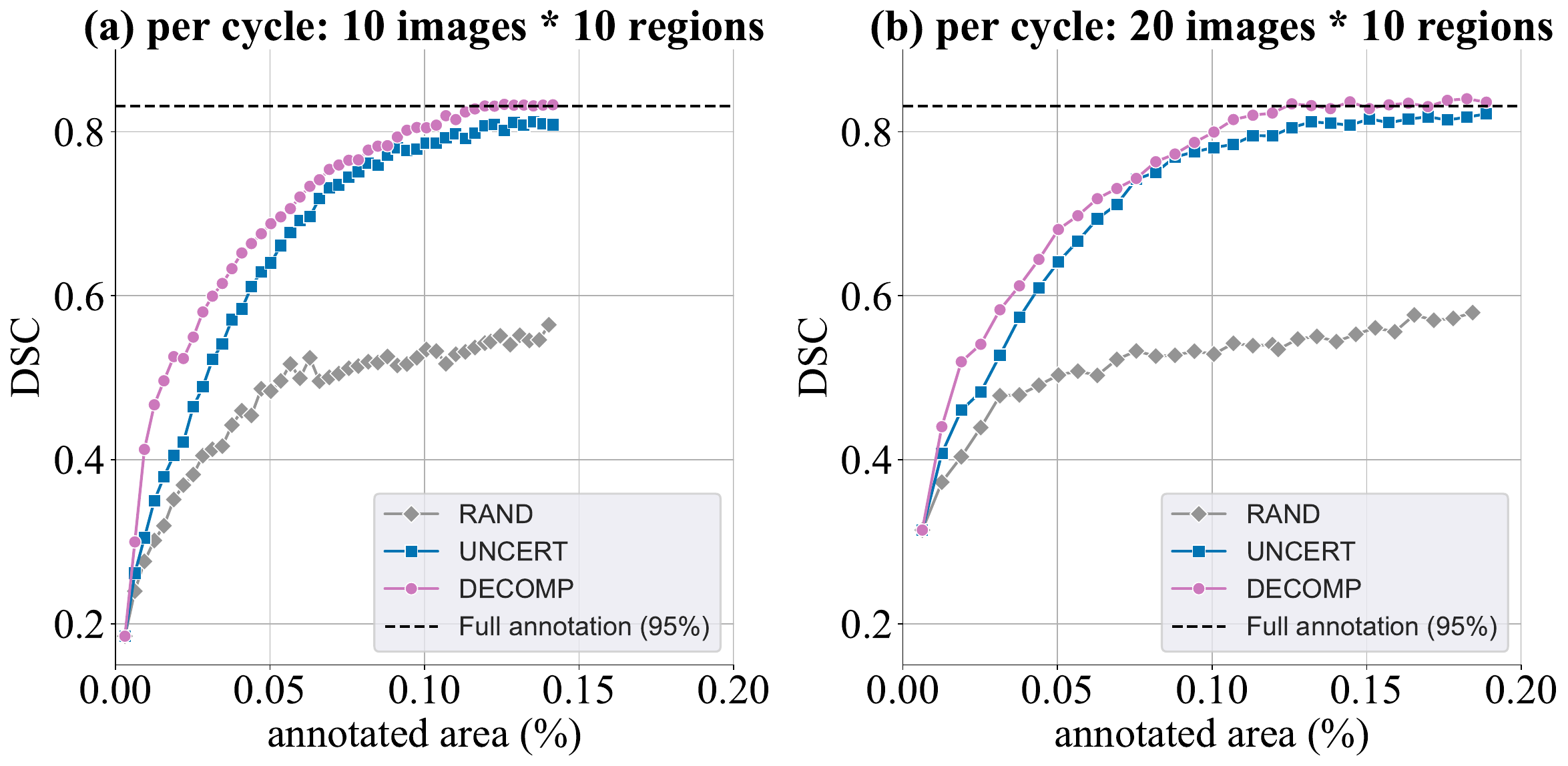}}}
\caption{Results on KiTS23 across $45$ and $30$ cycles for per-cycle annotation budget combinations $n_\text{image}\in\{10, 20\}$ and $n_\text{region}=10$, respectively. DSC as a function of annotated area (\%) for different methods. Means over five-fold cross validation.}
\label{fig:KiTS23_results}
\end{figure}
\noindent\textbf{Results on KiTS23 3-D segmentation} Evaluation on KiTS23 is shown in~\cref{fig:KiTS23_results}. The goal is to reduce the time-consuming voxel-level annotation. Results are consistent with those from the 2-D segmentation task on Cityscapes: UNCERT substantially outperforms RAND, possibly due to the small foreground ratio (\ie the kidney comprises only a small fraction of the abdominal CT scan). DECOMP further improves upon UNCERT, reaching $95\%$ of full-annotation performance with less than $0.15\%$ of the dataset volume annotated ($4,800$ regions of $32\times32$ pixels out of $386$ volumes of median size $177\times217\times217$ voxels). 

DIVERS(cluster), DIVERS(core-set), and BADGE are excluded from this experiment because the backbone requires fixed $128\times128\times128$ voxel inputs. Padding a $32\times32$ region to match this input size collapses it to a single voxel in the bottleneck feature map ($4\times4\times4$), making region scoring unreliable. This architectural constraint makes meaningful region selection infeasible without either redesigning the backbone or enlarging the annotation region (which easily introduces less informative areas). In contrast, DECOMP does not depend on region-level features; it requires only a single forward pass over the image and therefore remains directly applicable across different settings, including variations in backbone design and annotation region shape.

\subsection{Ablation Studies}

\begin{figure}[t]
{\centering\centerline{\includegraphics[width=1\columnwidth]{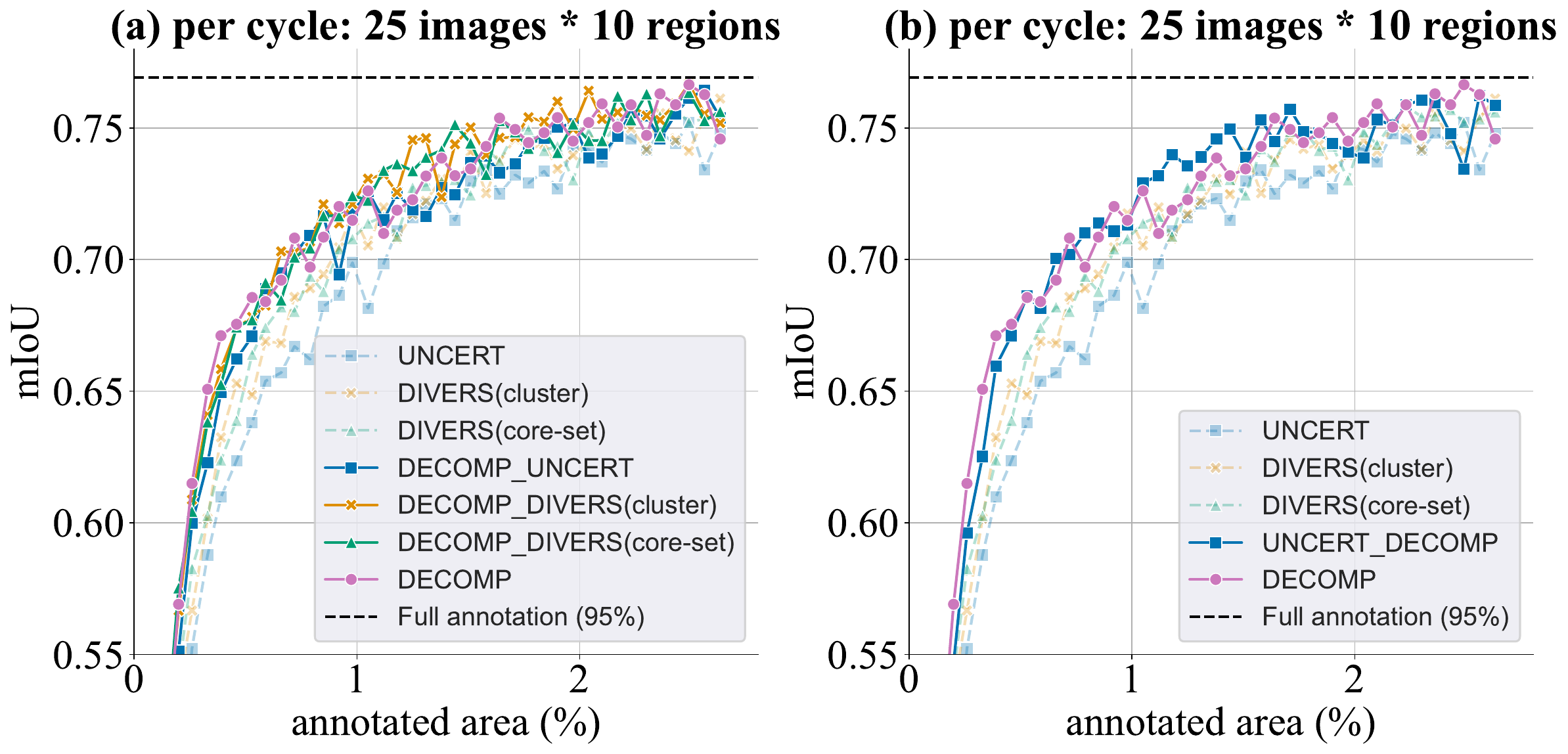}}}
\caption{Left: Effect of DECOMP image selection, replacing uncertainty image selection in UNCERT, DIVERS(cluster), and DIVERS(core-set) consistently improves performance (compare solid–dashed pairs). Right: Effect of DECOMP region selection, replacing UNCERT’s region selector with DECOMP's yields performance gains (UNCERT\_DECOMP vs.\ UNCERT). (Cityscapes)}

\label{fig:ablation_study}
\end{figure}

\noindent\textbf{Effect of DECOMP image selection.}
DIVERS(cluster) and DIVERS(core-set) rely on uncertainty-based image selection~\cite{yang2017suggestive,jin2021reducing}. DECOMP instead prioritizes images with diverse class predictions, especially from classes with high sampling weights, promoting both uncertainty and diversity. Replacing uncertain image selection with DECOMP’s in UNCERT, DIVERS(cluster), and DIVERS(core-set) yields consistent gains on Cityscapes (\cref{fig:ablation_study}(a)), showing DECOMP identifies more informative images. 

\noindent\textbf{Effect of DECOMP region selection.} Replacing UNCERT’s region selection with DECOMP’s improves performance on Cityscapes (UNCERT\_DECOMP vs. UNCERT)(\cref{fig:ablation_study}(b)). Additionally, when image selection is fixed to uncertainty sampling, DECOMP’s region querying consistently outperforms other diversity-based selectors (UNCERT\_DECOMP vs. DIVERS(cluster/core-set)).

Interestingly, DECOMP’s image and region selection strategies are each effective on their own, though using both does not always yield large additional gains. One possible reason is that their effects may overlap. Image selection promotes diversity by prioritizing images with multiple predicted classes, whereas region selection targets areas corresponding to different classes within each image. In datasets with inherently diverse images, such as Cityscapes, both strategies tend to identify similarly informative regions, limiting the incremental benefit of combining them. In contrast, on less diverse datasets like BRACS, where images often contain only one or a few adjacent lesion types, the combined use of DECOMP’s image and region selections offers clearer benefits, particularly under low annotation budgets (see Fig. 11 in the supplementary materials).

\begin{figure}[h]
  \centering
  \begin{minipage}[t]{0.475\columnwidth}
    \centering
    \includegraphics[width=\linewidth]{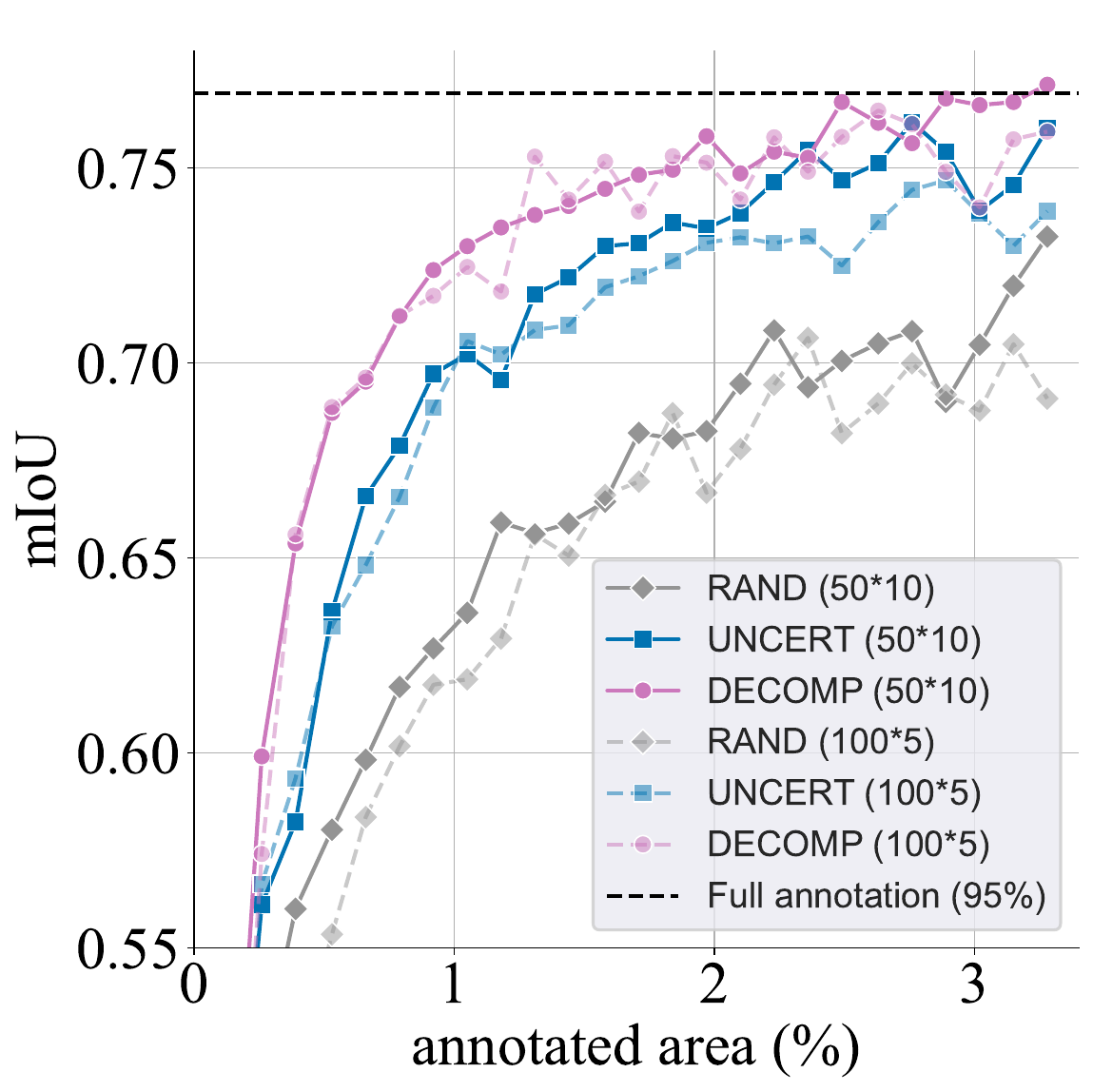}
    \caption{Dense annotation of fewer images vs. sparse annotation of more images. (Cityscapes)}
    \label{fig:image_selection}
  \end{minipage}
  \hfill
  \begin{minipage}[t]{0.48\columnwidth}
    \centering
    \includegraphics[width=\linewidth]{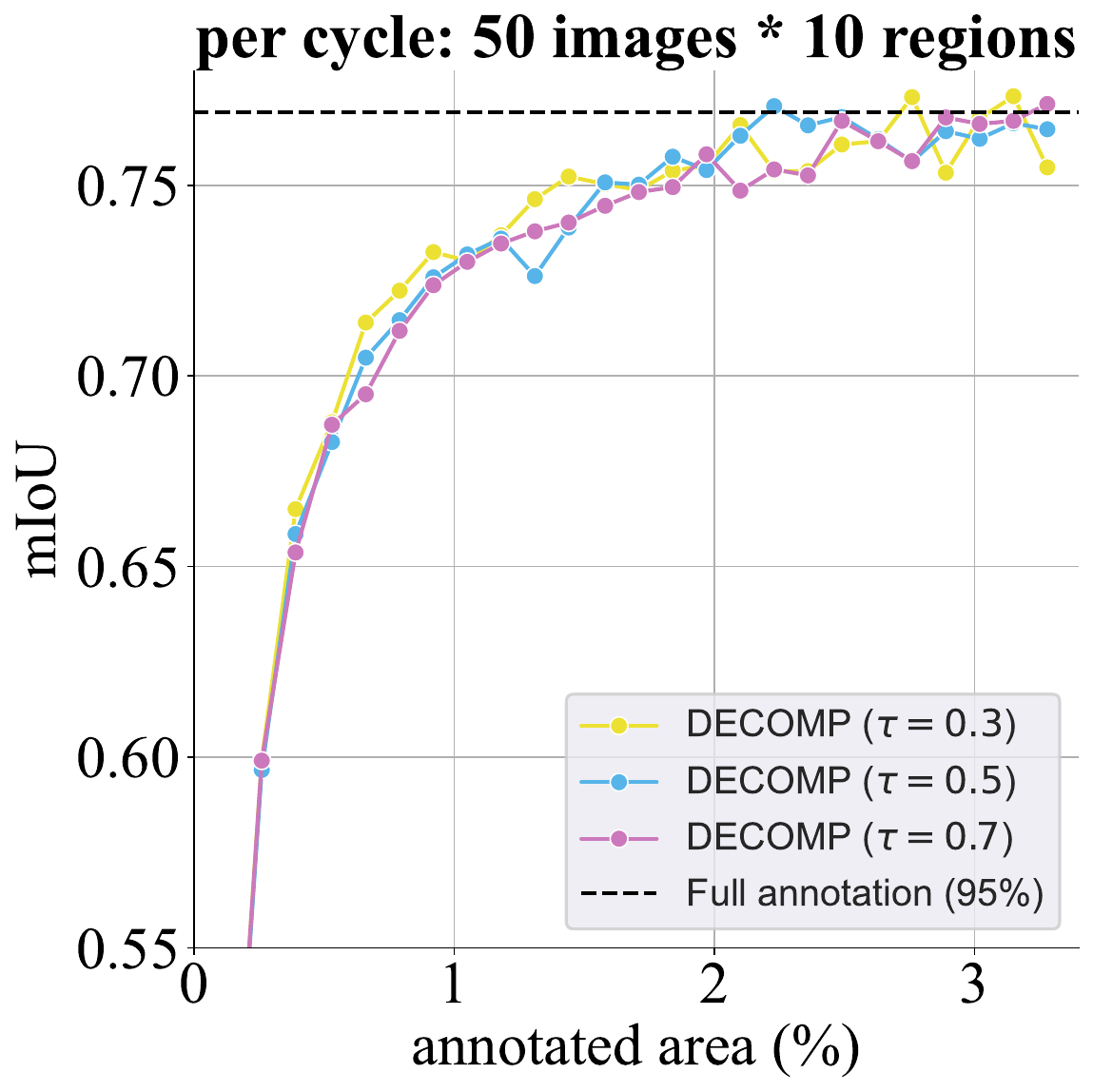}
    \caption{Impact of the confidence threshold $\tau$. (Cityscapes)}
    \label{fig:tau}
  \end{minipage}
\end{figure}

\noindent\textbf{Impact of annotation budget.}
A practical concern for real-world deployment of \ac{al} is the large number of annotation cycles often assumed in research settings, which may be unrealistic in clinical workflows. To investigate this, we explicitly evaluated DECOMP and comparison methods under different annotation budgets per cycle. Results on BRACS(~\cref{fig:BRACS_results}), Cityscapes(~\cref{fig:Cityscapes_results}) and KiTS23(~\cref{fig:KiTS23_results}) show that frequent updates with small per-cycle budgets do not consistently improve performance and may even reduce efficiency by limiting annotation diversity (\eg on BRACS for UNCERT). DECOMP remains robust across a wide range of budget settings, showing little sensitivity to this variation. This robustness, coupled with consistently outperforming all other methods, underscores its practicality for real-world applications.

\noindent\textbf{Dense vs. sparse annotation.}
Originally motivated by medical imaging applications to reduce cognitive loads, we propose restricting annotation regions to the most informative images rather than spreading them across all unlabeled images-the prior standard~\cite{mackowiak2018cereals}. We validate this by comparing dense annotation of fewer images ($n_\text{image}=50, n_\text{region}=10$) with sparse annotation of more images ($n_\text{image}=100, n_\text{region}=5$) under equal budgets. On Cityscapes (\cref{fig:image_selection}), UNCERT and RAND benefit from dense annotation, likely due to increased diversity and labeled context, while DECOMP remains robust.

\noindent\textbf{Hyperparameter robustness.}  
DECOMP has a single hyperparameter, the confidence threshold $\tau$ used to estimate class-wise confidence. To assess robustness, we compared $\tau\in \{0.3, 0.5, 0.7\}$. As shown in \cref{fig:tau}, a lower threshold ($0.3$) yields slightly better performance in early cycles, likely reflecting lower model confidence at that stage. However, the overall differences across all settings are small, indicating that DECOMP is largely insensitive to $\tau$ and does not require fine-tuning this parameter in practice.
\subsection{Qualitative Results and Further Analysis}

\begin{figure*}[h!]
\centering\centerline{\includegraphics[width=.82\paperwidth]{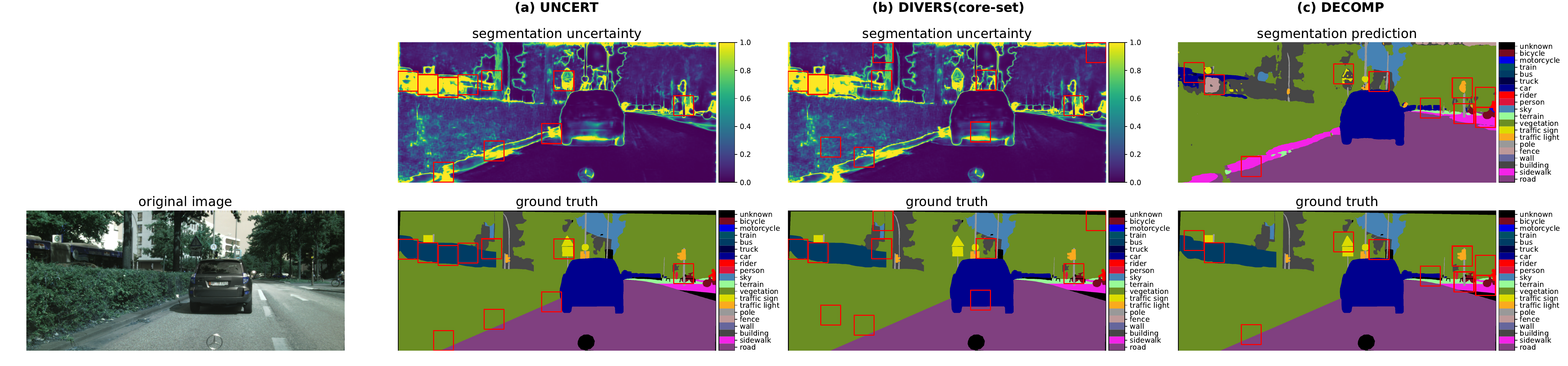}}
\caption{Example regions (red boxes) selected by different sampling methods on Cityscapes ($n_{\text{image}}=50, n_{\text{region}}=10$, $2^{nd}$ cycle).}
\label{fig:qualitative_results}
\end{figure*}

\begin{figure*}[h!]
{\centering\centerline{\includegraphics[width=.82\paperwidth]{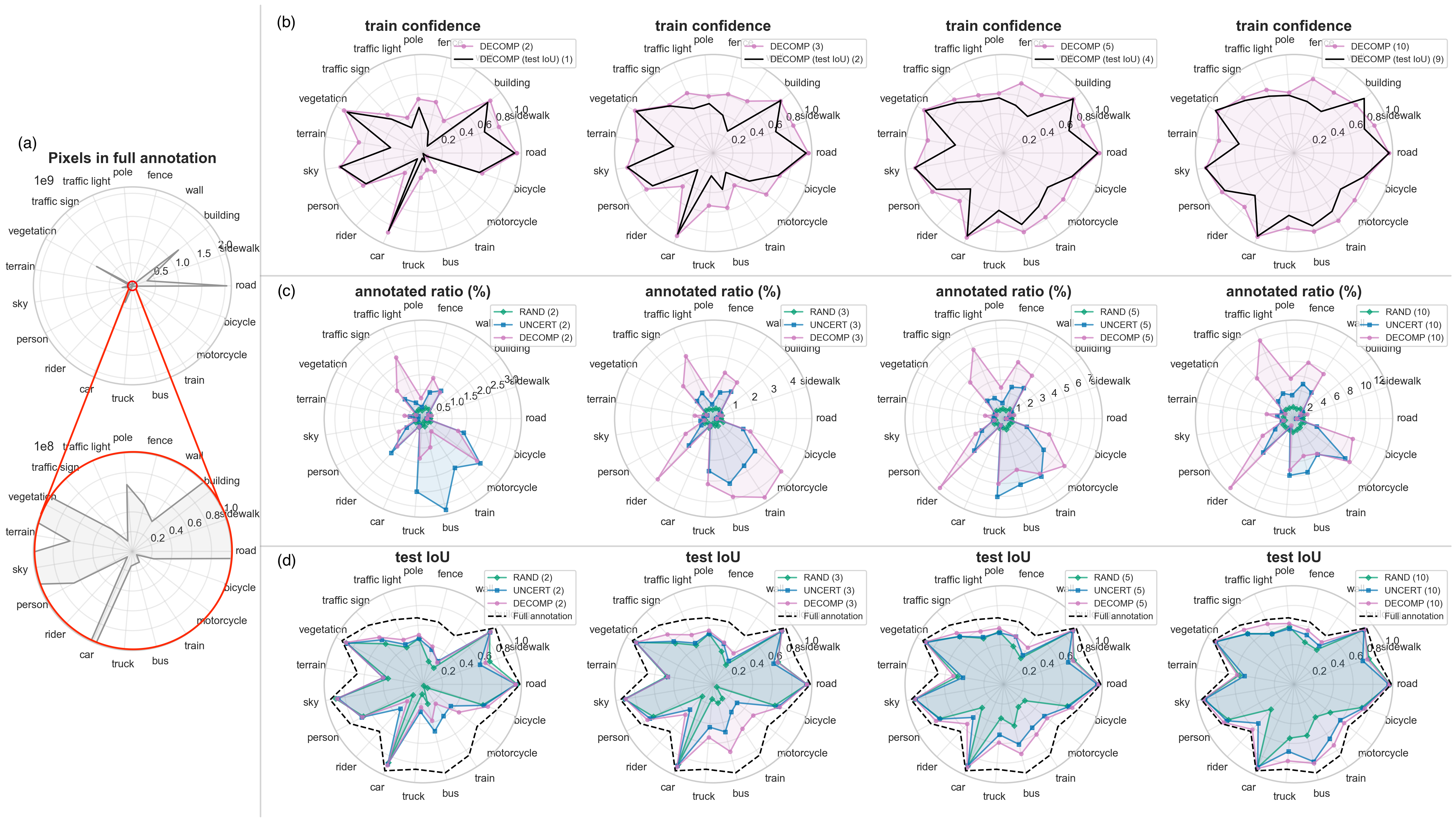}}}
\caption{(a) Class imbalance in the fully annotated Cityscapes dataset. (b–d) Results of cycles $2, 3, 5, 10$ ($n_\text{image}=100, n_\text{region}=5$), with each column showing one cycle (cycle number in parentheses): (b) class confidence rates estimated on the training set, (c) percentages of annotation area out of the full-annotation for each class selected by RAND, UNCERT, and DECOMP, and (d) model performance as IoU on the test set, the dashed line marks full-annotation performance. (b) additionally confirms the reliability of confidence estimates by comparing them to real test set performance using the same model $h^{(t-1)}$.}
\label{fig:Cityscapes_results_classwise}
\end{figure*}

\noindent\textbf{Qualitative results.} \Cref{fig:qualitative_results} shows regions selected by UNCERT, UNCERT(core-set), and DECOMP on Cityscapes. UNCERT concentrates on regions with high accumulated pixel uncertainty. UNCERT(core-set) instead chooses $n_\text{region}$ representative regions covering the feature space of the $3\times n_\text{region}$ most uncertain ones, yielding more diverse selections. For example, ``bus'' regions drop from four to two, ``vegetation–road'' transitions from three to one, while novel regions such as ``car'' and ``vegetation–building'' transition are added. DECOMP, leveraging segmentation class decomposition, captures an even broader range of classes, including ``traffic sign'', ``traffic light'', and ``rider''.

\noindent\textbf{Quality of pseudo-labels and class confidence estimations.}
A potential concern of DECOMP is that it depends on pseudo-labels for class confidence estimation and image decomposition, which may be noisy in early \ac{al} cycles and risk misguiding selection. To assess this, we compared our class-wise confidence estimates with true test performance (\cref{fig:Cityscapes_results_classwise} (b)). The close alignment, even in the earliest cycles, confirms that pseudo-labels provide a robust signal in uncertain classes identification. Moreover, noise in pseudo-labels is not purely detrimental: correct pseudo-labels help identify true annotation regions for the target class, while incorrect ones integrate model failures into annotations, which may be especially valuable for updating the current model (see example in~\cref{fig:qualitative_results} (c), where a selected region includes a bus misclassified as truck\&car, and another includes the transition of two bus sections misclassified as fence at the $2^{nd}$ \ac{al} cycle).

\noindent\textbf{Handling class imbalance.}  
We analyzed DECOMP’s handling of class imbalance by comparing the annotations it selected with those from other methods on Cityscapes. \Cref{fig:Cityscapes_results_classwise} (a) shows the extreme class imbalance of the dataset, while (c) shows the annotated ratios (\ie the fraction of each class’s true occurrence captured). For instance, the class ``traffic light'' is severely underrepresented and thus considered a minority class. An effective annotation sampling strategy is expected to detect its occurrence and query corresponding annotations. 

RAND simply follows dataset frequency, yielding similar annotation ratios across classes. UNCERT emphasizes poorly performing classes, such as ``truck'', ``bus'', ``train'', and ``motorcycle'', as reflected in the previous cycle’s test IoU (note that no test set information is used for annotation selection in any method). DECOMP also targets these classes, it selects less area than UNCERT on these classes but achieves higher performance, especially in later cycles, likely from added diversity. Notably, DECOMP captures more annotations for minority classes like ``traffic light'', ``traffic sign'' and ``rider'' via its decomposition step, boosting performance on these classes compared to RAND and UNCERT.

%% file: sec/5_conclusion.tex
\section{Conclusion, Limitations, and Future Work}

We introduced DECOMP, a new diversity-based \ac{al} strategy that leverages image decomposition to account for class balance and model weaknesses in annotation region selection. Requiring only a single forward pass per image, DECOMP is more efficient and easier to deploy than conventional feature-based methods. Across three datasets, it consistently outperforms strong baselines. Its robustness under both small per-cycle annotation budgets (favoring frequent model updates and avoiding unnecessary annotations) and large per-cycle annotation budgets (common in clinical workflows with limited annotator interaction) underscores its practical value.

An interesting finding is that DECOMP’s image and region selection offer strong independent performance but limited complementarity. Future work includes exploring alternative integrations of these two steps. Future work also involves adapting confidence thresholds across \ac{al} stages, measuring annotation effort via time rather than processed images or area, evaluating additional model backbones, and enabling re-selection of persistently uncertain regions for iterative refinement.

%% file: sec/7_acknowledgement.tex
\section*{Acknowledgements}

We acknowledge support by d.hip campus - Bavarian aim (J.Q.), the German Research Foundation (DFG) project 460333672 CRC1540 EBM, project 405969122 FOR2886 Pandora, projects 505539112, 520330054 and 545049923, as well as the scientific support and HPC resources provided by the Erlangen National High Performance Computing Center (NHR@FAU) of the Friedrich-Alexander-Universität Erlangen-Nürnberg (FAU). NHR funding is provided by federal and Bavarian state authorities. NHR@FAU hardware is partially funded by the German Research Foundation (DFG) – 440719683.

%% file: sec/6_supplementary.tex
\section{Datasets and Tasks}
\subsection{BRACS}
\textbf{BRACS}~\cite{brancati2022bracs} contains $320$ \ac{he}-stained \acp{wsi} (train/val/test: $193/68/59$). Annotating a \ac{wsi} requires complete \ac{roi} identification and classification. The full-annotation includes $3,163/602/626$ \acp{roi} across the splits, with $1$–$119$ \acp{roi} per slide (median: $8$). Notably, BRACS includes pre-cancerous lesions (\ac{adh} and \ac{fea}), which are clinically significant due to their progression risk~\cite{elmore2015diagnostic}. The classification task is to categorize a \ac{roi} into 7 classes (normal, benign, \ac{udh}, \ac{adh}, \ac{fea}, \ac{dcis}, and invasive carcinoma). Additionally, a 3-class task is performed to distinguish higher-level categories: benign (normal, benign, \ac{udh}), atypical tumor (\ac{adh}, \ac{fea}), and malignant tumor (\ac{dcis}, invasive). 7-class results are provided in supplementary materials. 

Annotation for the BRACS dataset followed a multi-step process~\cite{brancati2022bracs}. Three pathologists first determined the most aggressive tumor subtype in each \ac{wsi} as the image label. Then, each pathologist annotated a subset of \acp{wsi}, identifying and classifying \acp{roi}. While exhaustive identification of all regions was not required, especially for certain classes like normal tissue, efforts were made to maintain balanced class distributions. Consequently, the number of \acp{roi} per \ac{wsi} varies from $1$ to $119$ (median: $8$), with \acp{roi} of varying sizes to encapsulate entire diagnostic lesions. This annotation process further demonstrate the practical meaning of image selection that we proposed: minimizing the number of \acp{wsi} requiring expert review streamlines the real-world annotation workflow and increases opportunities for obtaining multi-annotator assessment.
\subsection{Cityscapes}
The Cityscapes dataset~\cite{cordts2016cityscapes} provides pixel-level urban scene segmentation for $2,975$ training and $500$ validation images across $19$ classes (road, sidewalk, building, wall, fence, pole, traffic light, traffic sign, vegetation, terrain, sky, person, rider, car, truck, bus, train, motorcycle and bicycle). The dataset exhibits extreme class imbalance in full-annotation: the minority class (motorcycle) accounts for only $0.27\,\%$ of the annotated pixels of the majority class (road). The task involves 2-D segmentation across the $19$ classes.

\subsection{KiTS23}
The KiTS23 dataset~\cite{heller2023kits21} includes $489$ 3-D abdominal CT images with manual annotations for kidneys, renal tumors, and renal cysts. The task involves hierarchical segmentation of the kidney region, the combined tumor and cyst regions, and the tumor alone. Accurate kidney tumor segmentation provides quantitative representations for risk stratification and treatment planning. Following~\cite{isensee2019attempt}, cases with confirmed or suspected faulty annotations (IDs: $23, 68, 125, 133, 15, 37$) were excluded. The CT scans contain $60$–$610$ slices (median: $177$). Dataset annotation involved identifying kidney regions in 3-D and delineating regions on axial planes~\cite{heller2023kits21}. 

\section{full-annotation Benchmarks, Implementation Details, and Evaluation Metrics}
\subsection{RoI Classification on BRACS}
We used HACT-Net~\cite{pati2022hierarchical} developed by the data provider, which constructs multi-level structural representations through cell- and tissue-graphs for breast cancer \ac{roi} subtyping. Nodes in the cell-graph represent nuclei, while tissue-graph nodes represent superpixel-based regions. For each node, features capturing morphological and spatial information are extracted to define inter-node interactions. The cell-graph and tissue-graph are processed independently by two distinct \acp{gnn}. The cell-\ac{gnn} processes a cell-node's feature and generates an embedding by aggregating information from neighboring nodes. The tissue-\ac{gnn} creates a tissue-node embedding by operating on its feature and the embeddings of spatially located cell-nodes within it. The final representation for the \ac{roi}, obtained by summing all tissue-node embeddings, is passed through a multi-layer perceptron (MLP) followed by a softmax layer to classify the \ac{roi} into a specific breast cancer subtype. The full-annotation performance for the 7-class and 3-class tasks was benchmarked on the test set as weighted F1 scores of $0.5540$ and $0.7320$, respectively (mean of five runs). \Ac{al} was used to annotate both training and validation set.
\subsection{2-D Segmentation on Cityscapes}
We used InternImage~\cite{wang2023internimage} with UperNet~\cite{xiao2018unified} as the decoder for benchmarking, owing to its top leaderboard performance\footnote{\url{https://www.cityscapes-dataset.com/benchmarks/}}. InternImage is an efficient \ac{cnn}-based foundation model featuring \acp{dcn}, which adaptively explore short- and long-range dependencies through learnable offsets and spatial aggregation scalars. It further splits each spatial aggregation process into multiple groups to learn diverse aggregation patterns for enhanced representation learning. We used InternImage-T for computational efficiency, given the iterative model training required in \ac{al} procedures. \Ac{miou} served as the evaluation metric. We trained maximally 64k iterations and stopped early when the validation performance stops improving for five consecutive 1k-iterations. The full-annotation performance achieved was $0.8096$ on the validation set (mean of five runs). \Ac{al} was used to annotate the training set.

\subsection{3-D Segmentation on KiTS23}
We used 3-D nnU-Net~\cite{isensee2021nnu} for benchmarking the full-annotation performance on KiTS23, due to its dominant usage among top KiTS21 leaderboard entries~\cite{heller2021state}. Class-averaged \ac{dice} score was used as the evaluation metric. We attained full-annotation performance of $0.8263, 0.8762, 0.8516$ for 2-D, 3-D-lowres and 3-D-fullres configurations, respectively (mean of five-fold cross-validation). We therefore used the 3-D-lowres configuration for all following experiments. Note that nnU-Net contains a preprocessing step of detecting foreground voxels using the full-annotation for dataset intensity normalization. Since the full-annotation is not available in \ac{al}, we detected the foreground voxels with intensity values in the range of $[-200, 500]$\,HU, resulting in a comparable \ac{dice} score of $0.8751$. Training on the fully annotated dataset with $1000$ epochs, as in~\cite{isensee2021nnu}, takes approximately seven hours using one NVIDIA A100 Tensor Core GPU. To improve computational efficiency, we reduced the training epochs to align with the total number of annotated regions in \ac{al} experiments, up to a maximum of $1000$ epochs. \Ac{al} was used to annotate the training set. \Ac{al} was used to annotate the training set for each corresponding cross-validation fold, and the average performance across the five folds was reported.

\section{Comparison Methods}
Here we provide more implementation details of comparison methods.

\noindent\textbf{RAND} : It randomly selects $n_{\text{image}}$ images and then choose $n_{\text{region}}$ non-overlapping regions at random locations within each image. Of note, foreground detection was performed before region selection on KiTS23. 

\noindent\textbf{UNCERT}: We compare to classic \textit{entropy}-based method for \ac{roi} classification and 2-D segmentation, where model outputs are processed with softmax, and to \textit{least confidence}-based method for the 3-D segmentation task with hierarchical classes, where model outputs are processed with sigmoid. These methods first select images with the highest average predictive uncertainty across all pixels (segmentation) or \acp{roi} (\ac{roi} classification), then choose the most uncertain regions within those images. For segmentation tasks, this involves splitting the image into overlapping regions with a stride of one pixel, calculating the average uncertainty of pixels within each region as the region uncertainty, and using non-maximum suppression to select $n_{\text{region}}$ non-overlapping regions with the highest uncertainties. For \ac{roi} classification, the region uncertainty is computed directly from its prediction. 

\noindent\textbf{DIVERS}: First, the $n_{\text{image}}$ most uncertain images are first selected. Then, $3\times n_{\text{region}}$ regions with the highest uncertainties are identified from each image to form a pool, from which $n_{\text{image}}*n_{\text{region}}$ regions are ultimately chosen using clustering (\textbf{DIVERS(cluster)}) or core-set identification (\textbf{DIVERS(core-set)}) based on region features.  This approach helps to avoid selecting similar regions across different images. Features from the bottleneck layer are used in segmentation tasks, while features from the penultimate layer before the classifier are used in \ac{roi} classification tasks. For the clustering-based method, we perform k-means clustering with the number of clusters set to $n_{\text{image}}*n_{\text{region}}$, and select the region with the highest uncertainty in each cluster, following~\cite{jin2021reducing}. For the core-set-based method, we followed~\cite{yang2017suggestive} to incrementally expand a set that maximally covers the latent space by iteratively including the region with the largest minimal distance to the already selected regions. 

\textbf{BADGE}~\cite{ash2019deep} selects informative samples by constructing gradient embeddings for each region, and applying clustering to promote both uncertainty and diversity. For \ac{roi} classification task, we follow the official implementation by weighting the feature of the penultimate layer by the discrepancy between predicted probabilities and one-hot pseudo-labels. For segmentation task, we replace with the feature obtained from the bottleneck layer, perform max pooling, and weight it with the averaged prediction discrepancy across all pixels in the region.

\section{Additional Results on BRACS}
\begin{figure}
{\centering\centerline{\includegraphics[width=1\columnwidth]{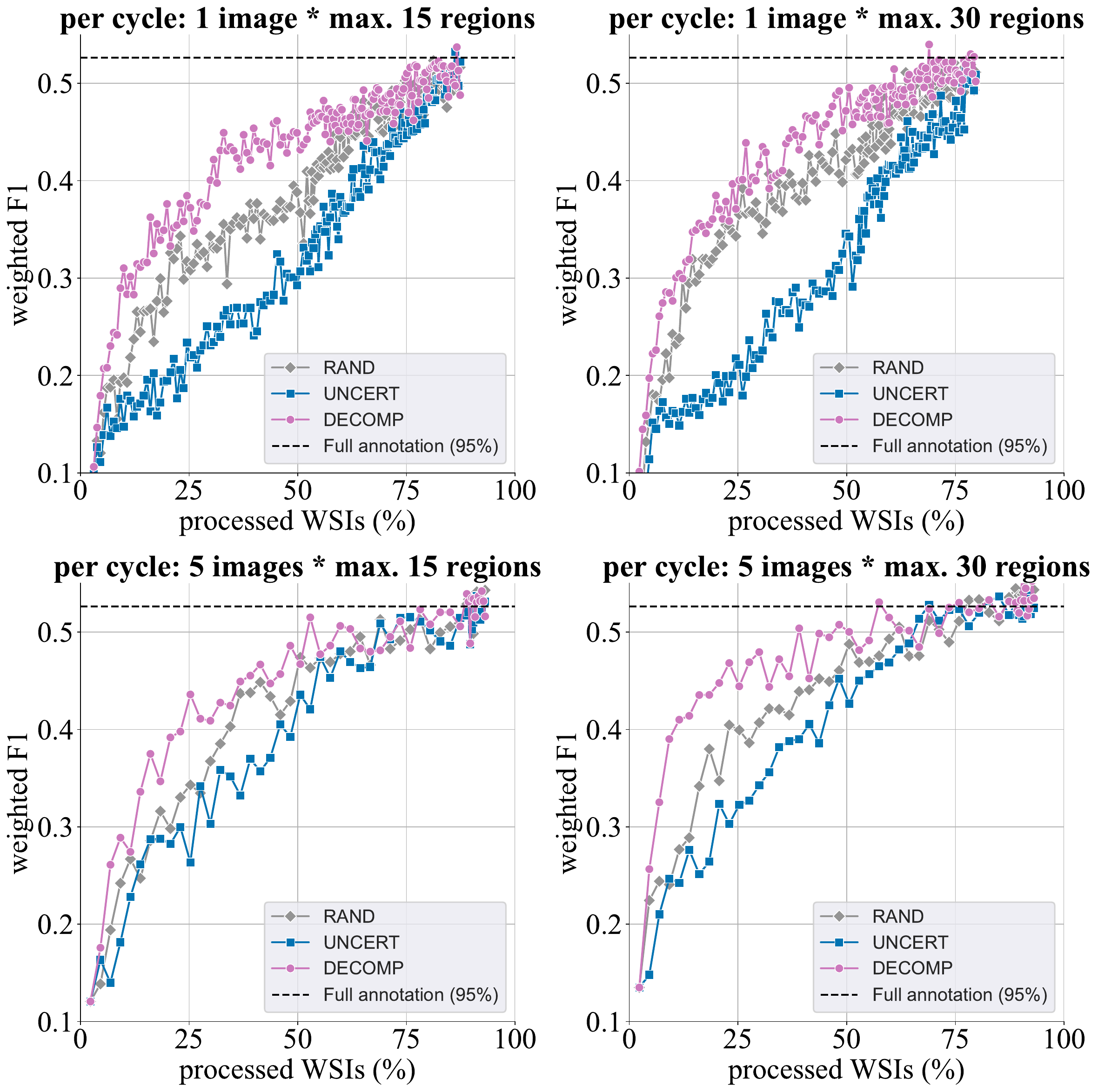}}}
\caption{Results on the BRACS dataset for the 7-class task across $160, 140, 50, 50$ cycles for \ac{al} hyperparameter combinations of $n_\text{image}\in\{1, 1, 5, 5\}$ and $n_\text{region}\in\{15, 15, 30, 30\}$, respectively. Weighted F1 as a function of annotated WSIs ($\%$) for different sampling methods.  Mean over five runs.}
\label{fig:BRACS_results_7_classes}
\end{figure}

\begin{figure}
{\centering\centerline{\includegraphics[width=1\columnwidth]{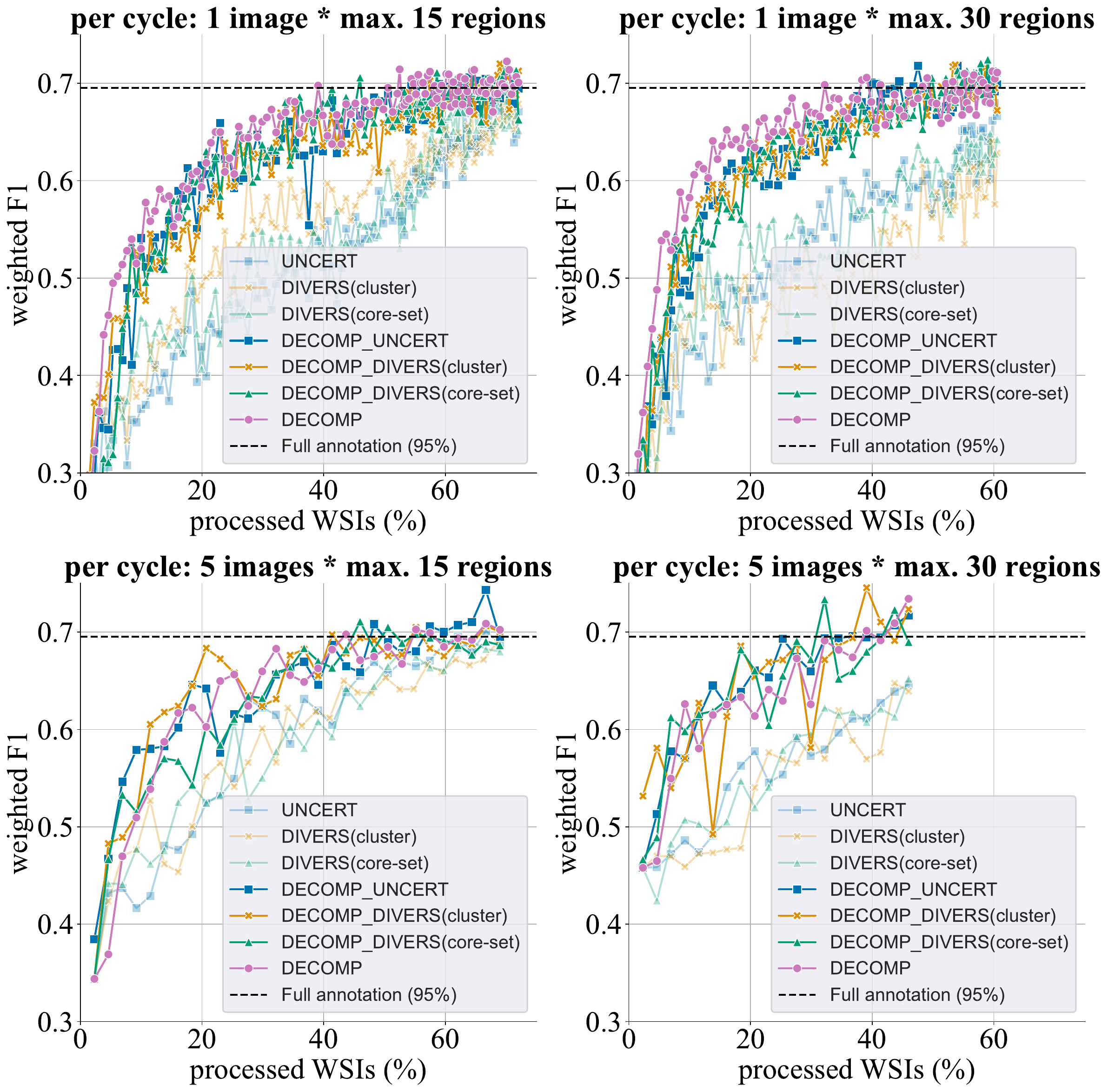}}}
\caption{Effect of image selection in DECOMP on the BRACS dataset.}  
\label{fig:ablation_study_BRACS}
\end{figure}


Due to page limits, we report only the 3-class task in the main text, with 7-class results in~\cref{fig:BRACS_results_7_classes} in the supplementary materials. The 7-class task is harder because of strong ambiguities among neighboring classes with shared morphology~\cite{brancati2022bracs}, resulting in lower full-annotation performance. Still, DECOMP maintains a clear sampling efficiency advantage over UNCERT and RAND.

\Cref{fig:ablation_study_BRACS} complements the Cityscapes ablation by showing the effect of DECOMP's image selection on BRACS. Similar to Cityscapes, switching image selection in UNCERT, DIVERS(cluster), and DIVERS(core-set) to DECOMP yields large gains, across all annotation budget settings. 